\documentclass{article} 
\usepackage{nips13submit_e,times}
\usepackage{hyperref}
\usepackage{url}
\usepackage{graphicx}
\usepackage{enumitem}
\usepackage{booktabs}

\title{Distinction between features extracted using \\Deep Belief Networks}

\author{
Mohammad Pezeshki \\
Department of Computer Eng. and IT\\
Amirkabir University of Technology\\
Tehran, Iran\\
\texttt{m.pezeshki@aut.ac.ir} \\
\And
Sajjad Gholami \\
Department of Computer Eng. and IT\\
Amirkabir University of Technology\\
Tehran, Iran\\
\texttt{s.gholami@aut.ac.ir} \\
\AND
Ahmad Nickabadi \\
Department of Computer Eng. and IT\\
Amirkabir University of Technology\\
Tehran, Iran\\
\texttt{nickabadi@aut.ac.ir} \\
}

\nipsfinalcopy 

\begin{document}

\maketitle

\begin{abstract}
Data representation is an important pre-processing step in many
machine learning algorithms. There are a number of methods
used for this task such as Deep Belief Networks (DBNs) and Discrete
Fourier Transforms (DFTs). Since some of the features extracted using
automated feature extraction methods may not always be related to a
specific machine learning task, in this paper we propose two methods in
order to make a distinction between extracted features based on their
relevancy to the task. We applied these two methods to a Deep Belief
Network trained for a face recognition task.
\end{abstract}

\section{Introduction}

Efficiency of many machine learning algorithms depends on the quality of features
 used for training [1]. There are some automated 
feature extraction methods such as Principle Component Analysis and Deep Belief Networks. The result of these methods is potentially useful,
  but there is one issue with these features. It is not always transparent 
  which features will be relevant for a given machine learning task. As a result, it 
  would be a great job to separate extracted features based on their relevant to the task.

One of the state-of-the-art tools for feature extraction is Deep Belief Network (DBN). 
 It would be useful if we were able to distinguish between 
 nodes which present different features. For example, in a face recognition 
 task, the main subject of the task is objects of face and side information such are considered as noise. If we use a DBN for 
 feature extraction, it is expected that some nodes in the last layer of the DBN 
 present the face and others present side information. Therefore, if we find nodes
presenting face singly, obviously the efficiency of the face recognition task would
be increased significantly. In this paper, we propose two methods in order to
make a distinction between last layer nodes of a DBN and in particular, examine 
the ability of a DBN to separate different features and represent them in 
distinct groups of nodes.

\section{Deep Belief Networks}

Deep Belief Networks (DBNs) are probabilistic graphical models which have multiple hidden layers. DBN is a mixed directed-undirected model such that all layer are connected with directed links except the top layer which forms an undirected bipartite graph. [Figure 1 (a)]. Hinton et al. introduced a fast greedy layer-wise algorithm which can be used for learning DBNs. [2]

DBNs can be constructed by staking multiple bipartite undirected graphical models called Restricted Boltzmann Machines (RBMs). RBM is a Boltzmann Machine which is restricted to have only one hidden layer and one visible layer and also have no visible-visible and hidden-hidden connections [3]. A graphical depiction of an RBM is shown in Figure 2 (b).

\begin{figure}[h]
\centering
\includegraphics[width=100mm]{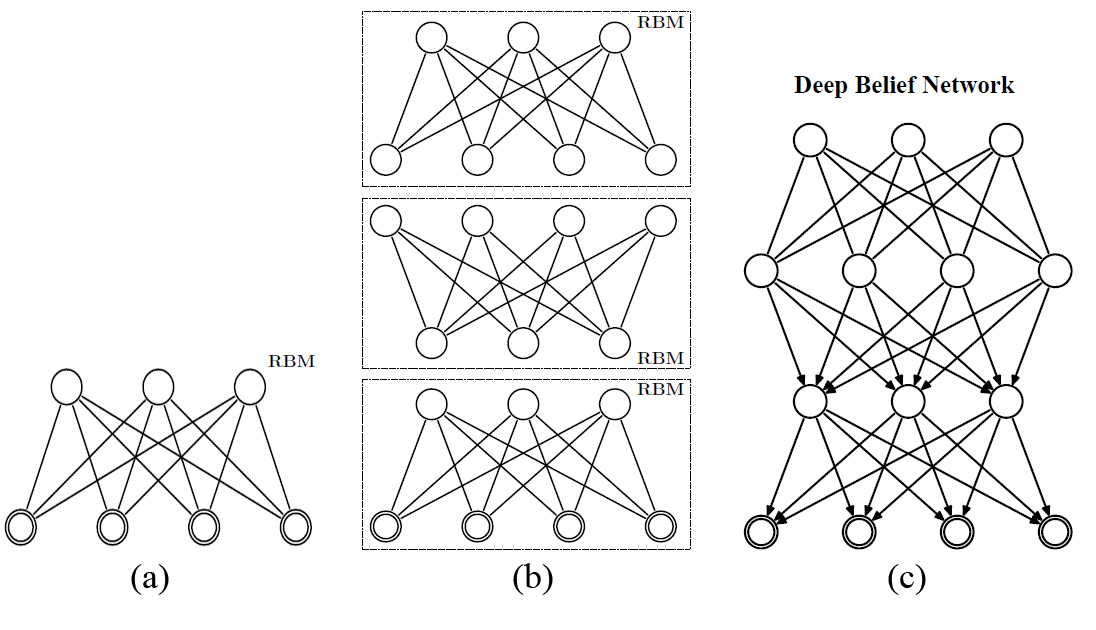}
\caption{(a) Restricted Boltzmann Machine. (b) A stack of RBMs. (c) The corresponding DBN. [4] }
\label{RBMpic}
\end{figure}

\section{Proposed Methods}

In this section, we will discuss proposed methods to make a distinction between
last layer nodes in a DBN.

\subsection{Method of Variances}

This method based upon the fact that inputs with different aspects (set of features) activate 
different nodes. Trying this process on some same-aspect inputs should force some nodes
to have a significant variation against others. 
If we feed the network with a group of inputs consisting of just one aspect,
 the values of some particular nodes in the last layer would
change significantly. Consequently, these nodes would have higher variations. Hence, a statistical
criterion such as Variance could be a good tool to distinguish between different
kinds of nodes.

\subsection{Method of Relative Activities}

The second method relies on the concept of relative activity. Relative activity is
an indicator for revealing the dependency of last layer nodes of a network to the
features of the given input. In this technique, relative activity of nodes can be
computed by subtracting the values of top layer nodes for two kinds of inputs.
First input consists of only one feature, and second input consists of previous
feature alongside another feature.

\section{Experimental results}
To evaluate the above-mentioned methods, we train a DBN with 4 hidden layers: 2000-1000-500-100. Training and testing done using the following dataset which consisted three parts:
      \begin{enumerate}
\item Face images from CMU PIE face database [5], size: 10,000
\item Handwritten digits from MNIST dataset [6], size: 5000
\item Face images corrupted by digit images, size: 5000
      \end{enumerate}
Some sample inputs are shown in Figure 2.
\begin{figure}[h]
\centering
\includegraphics[width=110mm]{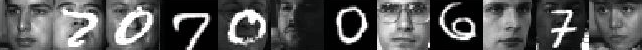}
\caption{Sample inputs.}
\label{sample}
\end{figure}
 
To discover the nodes presenting the face images we applied our two proposed
methods in the following ways:

\subsection{Using method of Variance}
According to method 1, a group of inputs consisting of faces
images singly are fed to the network. Now the Variance of nodes is computed and nodes with a Variance higher than 0.1
 are considered as nodes which present the face images. Again the DBN is fed with another input consisting of digit images. In
the same way, nodes with a Variance upper than 0.1 have a higher activity in
comparison with other nodes as shown in Figure 2-a.

\subsection{Using method of Relative activities}
According to method 2, each mixed image and its corresponding 
clear digit image are given to the network respectively. Node-from-node difference 
between last layer nodes for these two images show the relative activity. 
Finally, the average relative activity for all images are computed and nodes with an average relative activity higher than 0.7 are considered as nodes 
presenting the face features. This process is illustrated in Figure 2-b.

\begin{figure}[h]
\centering
\includegraphics[width=120mm]{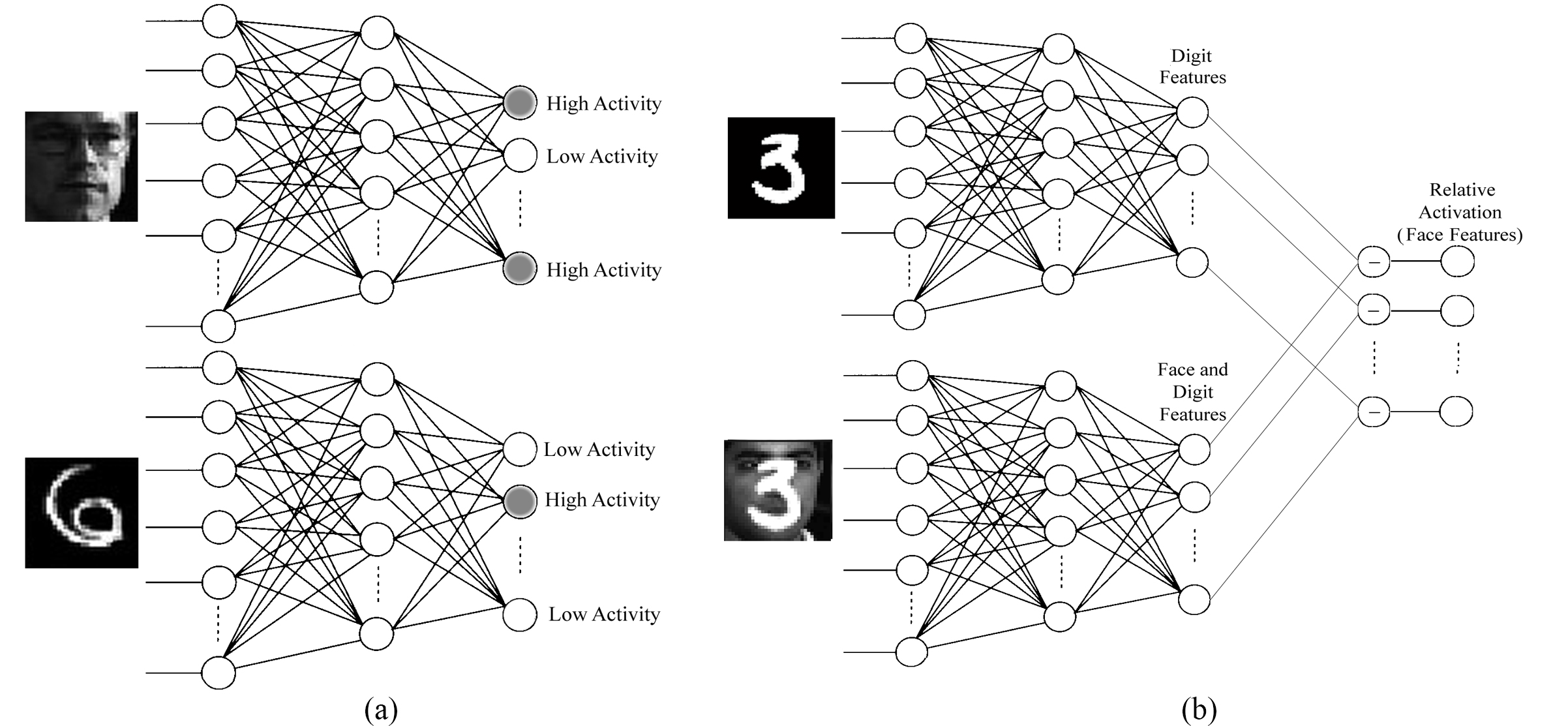}
\caption{{\bf (a)} Different images activates different nodes. (Method of Variances)
{\bf (b)} Relative activity can be computed by subtracting the values of nodes. (Method of Relative activities) }
\label{123}
\end{figure}

By applying methods mentioned in the preceding paragraphs, we discovered face
nodes (the nodes which present faces images). Now when a
mixed image is fed to the DBN, all nodes are active. To reconstruct the whole
face image which was previously corrupted by a digit, it is necessary to make
digit nodes (the nodes which present digit images) inactive. Digit nodes would
be inactive, when a neutral value is put instead of their current value. These
neutral values can be computed by averaging on the values of these nodes when
only face images are fed to the network.
Now only the face nodes are used in reconstruction process in practice. The
Figure 3 shows how the results of reconstruction process is improved when digit
nodes are inactivated.
\begin{figure}[h]
\centering
\includegraphics[width=95mm]{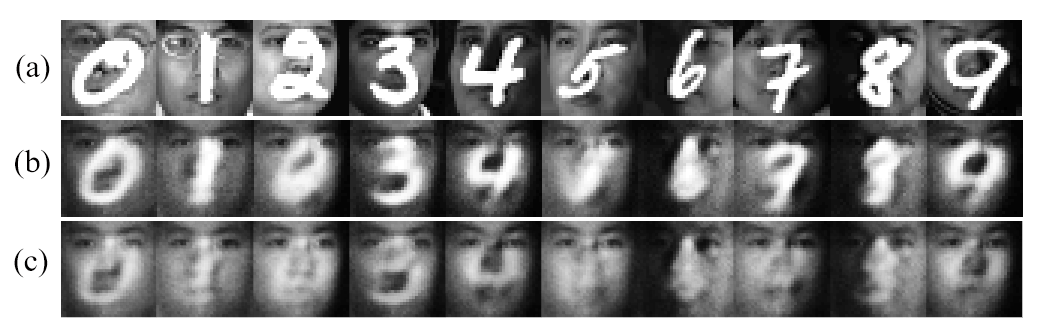}
\caption{{\bf (a)} Corrupted images which are fed to the network. {\bf (b)} 
Reconstructed images without changing the values of digit nodes. {\bf (c)} Reconstructed images when digit nodes are inactivated.}
\label{123}
\end{figure}

\section{Conclusion}
In this paper we focused on the properties of the features extracted using Deep
Belief Networks. Obviously, it would be quite useful if we are able
to make a distinction between the features extracted using a DBN. We proposed two novel methods in
order to understand which nodes are presenting which features. In our methods
Variance and Relative activity are two criteria to make a distinction between
nodes. We evaluated these methods on a data set consisting of MNIST handwritten 
digits and CMU PIE faces databases.

\subsubsection*{References}
\small{
[1] Bengio, Y., Courville, A. C., \& Vincent, P. Unsupervised feature learning and deep learning: A review and new perspectives. CoRR abs/1206.5538 (2012)

[2] Hinton, G. E., Osindero, S., \& Teh, Y. W. A fast learning algorithm for deep belief nets. Neural computation, 18(7), 1527-1554 (2006)

[3] Hinton, G. A practical guide to training restricted Boltzmann machines. Momentum, 9(1) (2010)

[4] Salakhutdinov, R. Learning deep generative models (Doctoral dissertation, University of Toronto) (2009)

[5] Sim, T., Baker, S., \& Bsat, M. The CMU pose, illumination, and expression database. Pattern Analysis and Machine Intelligence, IEEE Transactions on, 25(12), 1615-1618 (2003)

[6] LeCun, Y., \& Cortes, C. The MNIST database of handwritten digits (1998)
}

\end{document}